\PassOptionsToPackage{numbers, compress}{natbib}
\documentclass{article}
\usepackage[preprint]{neurips_2025}

\usepackage[utf8]{inputenc}
\usepackage[T1]{fontenc}
\usepackage{hyperref}
\usepackage{url}
\usepackage{booktabs}
\usepackage{amsmath, amssymb, amsfonts, bm}
\usepackage{nicefrac}
\usepackage{microtype}
\usepackage{xcolor}
\usepackage{graphicx}
\usepackage{subcaption}
\usepackage{wrapfig}
\usepackage{multirow}
\usepackage{makecell}
\usepackage{enumitem}
\usepackage{array}
\usepackage{colortbl}
\usepackage{pgfplots}
\pgfplotsset{compat=1.18}

\usepackage{algorithm}
\usepackage{algpseudocode}

\usepackage{tikz}
\newcommand*\circled[1]{\tikz[baseline=(char.base)]{
  \node[shape=circle,draw,inner sep=1pt] (char) {\footnotesize #1};}}

\usepackage{pifont}

\title{KerZOO: Kernel Function Informed Zeroth-Order Optimization for Accurate and Accelerated LLM Fine-Tuning}

\author{
  Zhendong Mi\textsuperscript{1}, 
  Qitao Tan\textsuperscript{2}, 
  Xiaodong Yu\textsuperscript{1}, 
  Zining Zhu\textsuperscript{1}, 
  \textbf{Geng Yuan}\textsuperscript{2}, 
  \textbf{Shaoyi Huang}\textsuperscript{1}\\
  \textsuperscript{1}Stevens Institute of Technology \quad
  \textsuperscript{2}University of Georgia \quad \\
  \texttt{\small \{zmi2, xyu38, zzhu41, shuang59\}@stevens.edu}, 
   \texttt{\small \{qitaotan, geng.yuan\}@uga.edu}
}

\begin{document}

\maketitle

\begin{abstract}
  Large language models (LLMs) have demonstrated impressive capabilities across numerous NLP tasks. Nevertheless, conventional first-order fine-tuning techniques impose heavy memory demands, creating practical obstacles to real-world applications. Zeroth-order (ZO) optimization has recently emerged as a promising memory-efficient alternative, as it circumvents the need for backpropagation by estimating gradients solely through forward passes—making it particularly suitable for resource-limited environments.
Despite its efficiency, ZO optimization suffers from gradient estimation bias, which significantly hinders convergence speed. To address this, we analytically identify and characterize the lower-order bias introduced during ZO-based gradient estimation in LLM fine-tuning. Motivated by tools in mathematical physics, we introduce a kernel-function-based ZO framework aimed at mitigating this bias and improving optimization stability.
  KerZOO achieves comparable or superior performance to existing ZO baselines in both full-parameter and parameter-efficient fine-tuning settings of LLMs, while significantly reducing the number of iterations required to reach convergence. For example, KerZOO reduces total GPU training hours by as much as 74\% and 44\% on WSC and MultiRC datasets in fine-tuning OPT-2.7B model and can exceed the MeZO baseline by 2.9\% and 2.6\% in accuracy.
  We show that the kernel function is an effective avenue for reducing estimation bias in ZO methods.
\end{abstract}

\section{Introduction}


The fine-tuning of pre-trained large language models (LLMs) for downstream tasks has attracted growing interest \cite{hu2021lora, dettmers2023qlora, gu2021efficient}. As model sizes continue to grow, full parameter fine-tuning (FT) becomes increasingly memory intensive, presenting significant computational challenges \cite{tan2025harmony, zhao2024second}.
To address GPU memory constraints associated with full fine-tuning, researchers have proposed parameter-efficient fine-tuning (PEFT) methods \cite{hu2022lora, li2021prefix, dettmers2023qlora, zhao2024galore}. These approaches update only a small subset of (additional) parameters, substantially reducing computational and storage costs while maintaining performance comparable to that of fully fine-tuned models.

Adaptive first-order optimizers such as Adam and AdamW \cite{loshchilov2017decoupled, kingma2014adam} are widely used for fine-tuning large language models (LLMs). However, these optimizers still incur substantial memory consumption, primarily due to the backpropagation process required for gradient computation in first-order optimization. To address the limitations, zeroth-order (ZO) optimization has emerged as a promising memory-efficient paradigm for LLM fine-tuning, attracting significant attention~\cite{zhang2024revisiting, zhao2024second, malladi2023fine}. Through fine-tuning large language models with only forward pass, it can achieve substantial memory reductions, makes it feasible to train and store LLMs on consumer hardware without the need for backpropagation.

However, while zeroth-order optimization methods significantly reduce memory consumption, they achieve this at the cost of slower convergence and reduced accuracy. Compared to first-order methods, ZO methods exhibit markedly inferior performance in terms of both convergence speed and time (GPU hours)~\cite{tan2025harmony}. 
Previous studies have analyzed the underlying causes of this issue, attributing it primarily to two key factors: (1) The parameters of LLMs often exhibit heterogeneous curvature across different dimensions~\cite{sagun2016eigenvalues, ghorbani2019investigation, zhang2020adaptive}. Such significant variation in second-order derivatives can cause ZO methods to converge toward saddle points, 
substantially impeding overall optimization convergence.
(2) Zeroth-order optimization methods estimate gradients via randomly sampled perturbations~\cite{malladi2023fine, gautam2024variance}. Consequently, when the sampled directions are suboptimal, the induced lower-order bias in the gradient estimation can significantly hinder convergence speed and degrade overall accuracy~\cite{lobanov2023accelerated, akhavan2024gradient}.

The first issue mentioned above has been analyzed and partially addressed in HIZOO~\cite{zhao2024second}. 
In this work, we focus on the second issue of the lower-order bias introduced by random perturbations in zeroth-order optimization (ZO). 
To mitigate the issue, 
we propose KerZOO, a \underline{ker}nel function informed \underline{Z}eroth-\underline{O}rder \underline{O}ptimization approach, to mitigate the lower-order estimation bias, thus improving the 
convergence speed during LLMs fine-tuning.
Experimental results show that KerZOO can dramatically reduce the required training steps, making the optimization process more efficient without sacrificing accuracy compared with the baselines.
We summarize our contributions as follows:


\begin{itemize}[leftmargin=*]
    \item We provide theoretical analysis on how existing zeroth-order method can introduce lower-order bias in gradient estimation of LLMs fine-tuning.   
    \item For the first time, we show that the kernel function can help mitigate the lower-order bias issue in zeroth-order optimization for LLMs fine-tuning with theoretical analysis. Moreover, we provide the design principle of the kernel function, aiming to remove the lower-order bias in LLMs fine-tuning, thus improving the convergence speed.
    \item We conduct extensive experiments across different models including encoder-only model (e.g., RoBERTa-large) and autoregressive language models (e.g., OPT and LLaMA). Experimental results show that KerZOO can achieve high accuracy and faster convergence. For example, for OPT-2.7B model fine-tuning, we can achieve up to 2.9\% and 2.6\% higher accuracy with 74\% and 44\% less GPU hours in convergence compared with the baseline on WSC and MultiRC datasets, respectively.
\end{itemize}



\section{Preliminaries and Background}

\subsection{Zeroth-Order Optimization}



Based on~\cite{bychkov2024accelerated}, a zeroth-order estimator of the gradient of a smooth function $f: \mathbb{R}^d \to \mathbb{R}$ at $x \in \mathbb{R}^d$ without accessing its  gradient can be defined as:
\begin{equation}
    \hat{g} = \frac{f(x + \epsilon u) - f(x - \epsilon u)}{2\epsilon} u,
     \label{eq1}
\end{equation}

where $\hat{g}$ is the estimated gradient at $x$, $u \in \mathbb{R}^d$ is a random unit direction, typically sampled uniformly from the unit sphere $\mathbb{S}^{d-1}$ or from a standard Gaussian distribution, and $\epsilon$  is a small positive value denoting a small perturbation step size.
Taking the expectation of both sides of Equation~\ref{eq1}, 
we obtain:
\begin{equation}
\mathbb{E}[\hat{g}] = \nabla f(x) + O(\epsilon^2),
\end{equation}
where $\hat{g}$ is an approximately unbiased estimator of $\nabla f(x)$ when $\epsilon$ is small, and $\mathbb{E}[\hat{g}]$ is the averaged estimated gradient over multiple random perturbations.
%
%
%
%
%
%
Given the learning rate $\alpha$ and the parameters $x$ in the $t$-th iteration, we can use the averaged estimated gradient to update the parameters:
\begin{equation}
x_{t+1}=x_t-\alpha \mathbb{E}[\hat{g}]
\end{equation}

\subsection{Analysis of ZO optimization}


Zeroth-order optimization techniques~\cite{malladi2023fine,bach2016highly,tan2025harmony,zhao2024second} have been extensively studied recently to address the substantial memory consumption problem in first-order optimization~\cite{kingma2014adam,zhou2024towards,hu2021lora}, since it only needs two forward passes per step and eliminates the need for backpropagation.
Despite its effectiveness in reducing memory consumption in LLMs fine-tuning,
the convergence speed in LLMs fine-tuning is notably slower compared to first-order (FO) optimization methods. 
We summarize the reason from two aspects:
Firstly, the parameter space often exhibits heterogeneous curvature across different dimensions in LLMs fine-tuning, leading to convergence instability or even failure~\cite{zhao2024second}; Secondly, ZO methods rely on randomly sampled perturbation directions, thus an inappropriate choice of direction may introduce large estimation bias in the gradient, resulting in degrading both convergence speed and overall optimization performance~\cite{lobanov2023accelerated, akhavan2024gradient}. To address the second issue, we propose a kernel-based zeroth-order method designed to eliminate lower-order bias in gradient estimates, thereby achieving faster and more effective convergence.

\section{Proposed Method} \label{method}

\textbf{Motivation.} 
In existing works in zeroth-order optimization for LLMs fine-tuning, perturbations are directly applied to the to be optimized
variables, without considering the bias introduced by the perturbations
However, due to the inherent stochastic nature of zeroth-order methods, excessive bias can significantly slow down convergence and degrade optimization performance \cite{akhavan2024gradient,bach2016highly, ghadimi2013stochastic}. Taking the inspiration, we propose a zeroth-order optimization method incorporating kernel functions to mitigate the lower-order estimation bias,
thus improving convergence speed during LLMs fine-tuning.


\subsection{Revisiting Zeroth-order Optimization for LLMs}


\medskip

Given a large language model with parameters $\bm{\theta} \in \mathbb{R}^d$ and loss function $\mathcal{L}$, we can use ZO method to estimate the gradient on a minibatch $\mathcal{B}_t$ at the iteration step $t$ , based on the concepts of sampling and differencing, as shown below:
\begin{equation}
\nabla \mathcal{L}(\bm{\theta}_t; \mathcal{B}_t) = \frac{\mathcal{L}(\bm{\theta}_t + \epsilon \bm{u}; \mathcal{B}_t) - \mathcal{L}(\bm{\theta}_t - \epsilon \bm{u}; \mathcal{B}_t)}{2\epsilon} \bm{u} 
\label{eq4}
\end{equation}

where $\bm{u} \in \mathbb{R}^d$ and $\bm{u}$ is a unit vector sampled from unit Gaussian sphere (i.e., $u = \frac{v}{\|v\|},\ \text{where } v \sim \mathcal{N}(0, I_d)$), $\epsilon$ is the perturbation scale. We can also use multiple sampled $\bm{u}$ to get the $n$-average gradient:
\begin{equation}
\nabla \mathcal{L}(\bm{\theta}_t; \mathcal{B}_t) = \frac{1}{n}\sum_{i = 1}^{n} [\frac{\mathcal{L}(\bm{\theta}_t + \epsilon \bm{u_i}; \mathcal{B}_t) - \mathcal{L}(\bm{\theta}_t - \epsilon \bm{u_i}; \mathcal{B}_t)}{2\epsilon} \bm{u_i} ]
\end{equation}

Given the learning rate $\eta$ and the mini-batch data $\mathcal{B}_t$ at $t$-th iteration, once the estimated gradient $\nabla \mathcal{L}(\bm{\theta}; \mathcal{B}_t)$ is obtained, then ZO-SGD updates the parameters as follows:
\begin{equation}
    \bm{\theta}_{t+1} = \bm{\theta}_t - \eta \nabla \mathcal{L}(\bm{\theta}_t; \mathcal{B}_t)
    \label{eq:zo_sgd_update}
\end{equation}

\subsection{Problem in Existing Zeroth-order Optimization for LLMs}

Since ZO methods estimate gradients based on randomly sampled perturbations $\bm{u}$, suboptimal directions can introduce substantial lower-order bias in the gradient estimates, thereby reducing estimation accuracy and hindering convergence speed~\cite{lobanov2023accelerated, akhavan2024gradient}. To further analyze the issue,
we apply Taylor expansion on different terms in equation \ref{eq4}. On the same minibatch, we have:
\begin{equation}
\mathcal{L}(\bm{\theta} + \epsilon \bm{u}) = \mathcal{L}(\bm{\mathcal{\theta}}) + \epsilon  \langle \nabla \mathcal{L}(\bm{\mathcal{\theta}}), \bm{u} \rangle + \frac{(\epsilon )^2}{2} \bm{u}^\top \nabla^2 f(x) \bm{u} + \frac{(\epsilon )^3}{6} D^3 \mathcal{L}(\bm{\mathcal{\theta}})[\bm{u},\bm{u},\bm{u}] + O(\epsilon^4)
\end{equation}
\begin{equation}
\mathcal{L}(\bm{\theta} - \epsilon \bm{u}) = \mathcal{L}(\bm{\mathcal{\theta}}) - \epsilon  \langle \nabla \mathcal{L}(\bm{\mathcal{\theta}}), \bm{u} \rangle + \frac{(\epsilon )^2}{2} \bm{u}^\top \nabla^2 f(x) \bm{u} - \frac{(\epsilon )^3}{6} D^3 \mathcal{L}(\bm{\mathcal{\theta}})[\bm{u},\bm{u},\bm{u}] + O(\epsilon^4)
\end{equation}

where $\nabla \mathcal{L}(\bm{\mathcal{\theta}})$ is the gradient at $\mathcal{\bm{\theta}}$,
$\nabla^2 \mathcal{L}(\bm{\mathcal{\theta}})$ is the Hessian matrix at $\mathcal{\bm{\theta}}$,
and $D^3 \mathcal{L}(\bm{\mathcal{\theta}})[\bm{u},\bm{u},\bm{u}]$ denotes the third-order directional derivative of $\mathcal{L}$ along $\bm{u}$ three times. Taking the difference, we have:
\begin{equation}
\mathcal{L}(\bm{\theta} + \epsilon \bm{u}) - \mathcal{L}(\bm{\theta} - \epsilon \bm{u})= 2 \epsilon  \langle \nabla \mathcal{L}(\bm{\mathcal{\theta}}), \bm{u} \rangle + \frac{(\epsilon)^3}{3} D^3 \nabla \mathcal{L}(\bm{\mathcal{\theta}})[\bm{u},\bm{u},\bm{u}] + O(\epsilon^4)
\end{equation}

Therefore, the equation \ref{eq4} can be expressed as:
\begin{equation}
\frac{\mathcal{L}(\bm{\theta} + \epsilon \bm{u}) - \mathcal{L}(\bm{\theta} - \epsilon \bm{u})}{2\epsilon}\cdot \bm{u}=   \langle \nabla \mathcal{L}(\bm{\mathcal{\theta}}), \bm{u} \rangle \bm{u}+ \frac{(\epsilon)^2}{6} D^3 \nabla \mathcal{L}(\bm{\mathcal{\theta}})[\bm{u},\bm{u},\bm{u}]\cdot \bm{u}+ O(\epsilon^4)
\end{equation}

Taking the expectation (noting that $\mathbb{E}[\bm{u}\bm{u}^\top] = \frac{1}{d} I_d$), we have:
\begin{equation}
\mathbb{E}[\frac{\mathcal{L}(\bm{\theta} + \epsilon \bm{u}) - \mathcal{L}(\bm{\theta} - \epsilon \bm{u})}{2\epsilon}\cdot \bm{u}]= \frac{1}{d}   \nabla \mathcal{L}(\bm{\mathcal{\theta}})+ \mathbb{E}[\frac{(\epsilon)^2}{6} D^3 \nabla \mathcal{L}(\bm{\mathcal{\theta}})[\bm{u},\bm{u},\bm{u}]\cdot \bm{u}]+ O(\epsilon^4)
\end{equation}

As shown in the above formulation, zeroth-order methods yield estimated gradients containing higher-order bias terms $\mathbb{E}[\frac{(\epsilon)^2}{6} D^3 \nabla \mathcal{L}(\bm{\mathcal{\theta}})[\bm{u},\bm{u},\bm{u}]\cdot \bm{u}]+ O(\epsilon^4)$,
which are highly sensitive to the choice of perturbation direction $\bm{u}$, 
and a suboptimal sampling of $\bm{u}$ can result in large bias, leading to subpotimal gradient estimation and slow convergence.


\subsection{Kernel Function Informed Zeroth-Order Optimization}

Motivated by kernel smoothing techniques in mathematical physics, which are widely used to reduce estimation bias~\cite{nesterov2017random, bach2016highly,akhavan2024gradient}, we propose a kernel function informed zeroth-order optimization method for LLMs fine-tuning to address the challenge 
in gradient estimation arising from the high dimensionality of LLMs in which the randomly sampled perturbations
exhibit different values in different directions.


To control the perturbation magnitude, we incorporate a random scalar variable $r$ in equation \ref{eq4} and use $\hat{g}$ to denote the estimated gradient as:
\begin{equation}
\hat{g} = \frac{\mathcal{L}(\bm{\theta} + \epsilon r \bm{u}) - \mathcal{L}(\bm{\theta} - \epsilon r \bm{u})}{2 \epsilon} \bm{u}
\label{eq12}
\end{equation}
where $\bm{u}$ is a random unit vector (e.g., uniformly sampled from the unit Gaussian sphere),
and $\epsilon > 0$ is a small step size denoted as perturbation scale.

Assuming loss function $\mathcal{L}$ is at least third-order differentiable, by performing a Taylor expansion on $\mathcal{L}(\bm{\theta} \pm \epsilon r \bm{u})$, we have:
\begin{equation}
\mathcal{L}(\bm{\theta} + \epsilon r\bm{u}) = \mathcal{L}(\bm{\mathcal{\theta}}) + \epsilon  r\langle \nabla \mathcal{L}(\bm{\mathcal{\theta}}), \bm{u} \rangle + \frac{(\epsilon r)^2}{2} \bm{u}^\top \nabla^2 f(x) \bm{u} + \frac{(\epsilon r )^3}{6} D^3 \mathcal{L}(\bm{\mathcal{\theta}})[\bm{u},\bm{u},\bm{u}] + O(\epsilon^4)
\end{equation}
\begin{equation}
\mathcal{L}(\bm{\theta} - \epsilon r\bm{u}) = \mathcal{L}(\bm{\mathcal{\theta}}) - \epsilon r \langle \nabla \mathcal{L}(\bm{\mathcal{\theta}}), \bm{u} \rangle + \frac{(\epsilon r)^2}{2} \bm{u}^\top \nabla^2 f(x) \bm{u} - \frac{(\epsilon r)^3}{6} D^3 \mathcal{L}(\bm{\mathcal{\theta}})[\bm{u},\bm{u},\bm{u}] + O(\epsilon^4)
\end{equation}

Taking the difference between the two expressions, we get:
\begin{equation}
\mathcal{L}(\bm{\theta} + \epsilon r \bm{u}) - \mathcal{L}(\bm{\theta} - \epsilon r \bm{u})= 2 \epsilon r \langle \nabla \mathcal{L}(\bm{\mathcal{\theta}}), \bm{u} \rangle + \frac{(\epsilon r)^3}{3} D^3 \nabla \mathcal{L}(\bm{\mathcal{\theta}})[\bm{u},\bm{u},\bm{u}] + O(\epsilon^4)
\label{eq15}
\end{equation}
Combining Equation \ref{eq15} and Equation \ref{eq12}, we have:
\begin{equation}
\hat{g} = \frac{\mathcal{L}(\bm{\theta} + \epsilon r\bm{u}) - \mathcal{L}(\bm{\theta} - \epsilon r\bm{u})}{2\epsilon}\cdot \bm{u}=  r \langle \nabla \mathcal{L}(\bm{\mathcal{\theta}}), \bm{u} \rangle \bm{u}+ \frac{(\epsilon)^2 r^3}{6} D^3 \nabla \mathcal{L}(\bm{\mathcal{\theta}})[\bm{u},\bm{u},\bm{u}]\cdot \bm{u}+ O(\epsilon^4)
\end{equation}
where
the first term $r \langle \nabla f(x), u \rangle u$ can be used to approximate the true gradient $\mathcal{L}(\bm{\theta})$, and the second term $\frac{(\epsilon)^2 r^3}{6} D^3 \nabla \mathcal{L}(\bm{\mathcal{\theta}})[\bm{u},\bm{u},\bm{u}]\cdot \bm{u}$ introduces a leading bias of order $O(\epsilon^2)$.
%
To reduce the second-order bias, we introduce a kernel function $K(r)$, and modify the gradient estimator as:
\begin{equation}
\hat{g}_K = \frac{\mathcal{L}(\bm{\theta} + \epsilon r \bm{u}) - \mathcal{L}(\bm{\theta} - \epsilon r \bm{u})}{2\epsilon} K(r) \bm{u}
\end{equation}
Applying the Taylor expansions, we have:
\begin{equation}
\hat{g}_K = (r \langle \nabla \mathcal{L}(\bm{\mathcal{\theta}}), \bm{u} \rangle K(r)) \bm{u}+ \frac{(\epsilon)^2 r^3}{6} D^3 \nabla \mathcal{L}(\bm{\mathcal{\theta}})[\bm{u},\bm{u},\bm{u}]K(r) \bm{u}+ O(\epsilon^4)
\label{eq18}
\end{equation}
Taking the expectation of both sides of Equation~\ref{eq18}, we have:
\begin{equation}
\mathbb{E}[\hat{g}_K] = \mathbb{E}[rK(r)] \frac{1}{d} \nabla \mathcal{L} (\bm{\mathcal{\theta}})+ \mathbb{E} [r^3K(r)]\mathbb{E}[\frac{(\epsilon)^2 }{6} D^3 \nabla \mathcal{L}(\bm{\mathcal{\theta}})[\bm{u},\bm{u},\bm{u}] \bm{u}]+ O(\epsilon^4)
\label{eq19}
\end{equation}

To remove the lower-order bias (i.e., the second term in Equation~\ref{eq19}), we develop the following kernel function $K(r)$ design principle:

\begin{algorithm}
\caption{KerZOO}
\begin{algorithmic}[1]
\State \textbf{Inputs:} Starting points of the LLM parameters $\bm{\theta}_0^{ag} = \bm{\theta}_0  \in \mathbb{R}^d$ ($\bm{\theta}^{ag}_0$ is an intermediate variable as same as the model parameters $\bm{\theta}_0$, and it is also updated at each iteration), number of iterations $N$, perturbation number $n$, perturbation scale $\epsilon > 0$, kernel $K(\cdot)$, learning rate $\eta$, iteration constant $\beta_0=1$, gradient clip constant $R$.
\For{$t = 0$ to $N-1$}
    \State $\beta_t = 1 + \frac{t}{6}$
    \State $\bm{\theta}_t^{md} = \beta_t^{-1} \bm{\theta}_t + (1 - \beta_t^{-1}) \bm{\theta}_t^{ag}$
    \State Sample random perturbation $\bm{\bm{u}}_i\sim  \mathcal{N}(0, I_d)$, $r_i \sim \text{Uniform}[-1, 1]$ for $i = 1, \ldots, n$
    \State Compute batched gradient approximation:
    \[
    \bm{g}_K^t=\mathbb{E} [\hat{g}_K^t]= \frac{1}{n} \sum_{i=1}^{n} \frac{\mathcal{L}(\bm{\theta}_t + \epsilon r_i \bm{u}_i) - \mathcal{L}(\bm{\theta}_t - \epsilon r_i \bm{u}_i)}{2\epsilon} K(r_i) \bm{u}_i
    \]
    \State $\bar{\bm{\theta}}_{t+1} = \bm{\theta}_t - \eta\bm{g}_K^t$
    \State $\bm{\theta}_{t+1} = \min\left\{ 1, \frac{R}{\|\bar{\bm{\theta}}_{t+1}\|} \right\} \bar{\bm{\theta}}_{t+1}$
    \State $\bm{\theta}_{t+1}^{ag} = \beta_t^{-1} \bm{\theta}_{t+1} + (1 - \beta_t^{-1}) \bm{\theta}_t^{ag}$
\EndFor
\State \textbf{return} $\bm{\theta}_N^{ag}$
\end{algorithmic}
\end{algorithm}

\begin{itemize}[leftmargin=*]
    \item \textbf{First-moment condition:} $\mathbb{E}[r K(r)] = C$ ($C$ is a constant), ensuring that the estimator remains approximately unbiased for the true gradient; 
    \item \textbf{Third-moment condition:} $\mathbb{E}[r^3 K(r)] = 0$, eliminating the leading second-order bias term.
\end{itemize}

Based on the kernel function design principle, Equation~\ref{eq19} becomes $\mathbb{E}[\hat{g}_K] = \frac{C}{d} \nabla \mathcal{L} (\bm{\mathcal{\theta}})+ O(\epsilon^4)$.




By carefully constructing $K(r)$ to satisfy these moment conditions (see Section~\ref{kernel} for details), we can effectively remove the second-order (lower-order) bias, leading to a more accurate gradient estimator which have only fourth-order (higher order) bias. 
In our practical application, we can limit the $r$ in a smaller range as the iteration step increases (see Appendix \ref{var} for details). Additionally, to satisfy the condition $\mathbb{E}[r^3 K(r)] = 0$ and $\mathbb{E}[r K(r)] = C$, it is essential to perform multiple perturbations so that the expectation becomes statistically meaningful. Our kernel function informed zeroth-order optimization can be found in Algorithm 1.

\subsection{Kernel Function Design} \label{kernel}

Here we describe the details of the kernel function design. According to \cite{polyak1990optimal}, we consider $r$ uniformly distributed in $[-1,1]$, 
then we may choose \( K_\beta(r) = C\cdot\sum_{m=0}^{\beta} p'_m(0) p_m(r) \). 
\begin{equation}
K(r) = C\cdot\sum_{m=0}^{\beta} p'_m(0) p_m(r)
\end{equation}
In the expression,
\begin{equation}
p_m(u)=\sqrt{2m+1}L_m(u)
\end{equation}
$L_m(u)$ denotes the $m$-th Legendre polynomial. $p'(0)$ denotes the derivative of the polynomial at the point $m=0$. $\beta$ represents the order of the polynomial, corresponding to the highest power $s$ that the expression $\mathbb{E}[r^sK(r)]=0$ can accommodate. $C$ represents a constant.



For example, we have the following values for \( \beta \in \{1, 3, 5\} \):
\begin{align*}
K_1(r)  &=C\cdot 3r \\
K_3(r)  &= C\cdot \frac{15}{4}r(5 - 7r^2) \\
K_5(r)  &=C\cdot \frac{195}{64}r(99r^4 - 126r^2 + 35)
\end{align*}
Taking $\beta=3$ as an example:
\begin{equation}
\mathbb{E}[rK_3(r)]=\int_{-1}^{1} \frac{15C}{4}r^2(5 - 7r^2) \cdot \rho(r) \, dr = C
\end{equation}
\begin{equation}
\mathbb{E}[r^3K_3(r)]=\int_{-1}^{1} \frac{195C}{64}r^4(5 - 7r^2)\cdot \rho(r) \, dr = 0
\end{equation}
$\rho(r)$ is the probability density function of the variable $r$ and $\rho(r)=\frac{1}{2}$ if $r\sim \mathcal{U}[-1,1]$. As to higher order kernel function such as $K_5(r)$, we can also have $\mathbb{E}[r^5K_5(r)]=\int_{-1}^{1} \frac{195C}{64}r^6(99r^4 - 126r^2 + 35) \, dr = 0$, removing more higer-order bias term. In our experiment, we choose to use $K_3(r)$ as our experimental kernel. 


\section{Experiments}

\subsection{Experimental Settings}

\paragraph{Models.}
We experiment with both masked language models and autoregressive models. For masked language modeling, we use RoBERTa-large~\cite{liu2019roberta}. For autoregressive models, we consider the OPT~\cite{zhang2022opt} and LLaMA~\cite{touvron2023llama} models. Model scales range from 355M to 6.7B parameters, including OPT-2.7B, OPT-6.7B, and LLaMA-3-3B and LLaMa-3-8B, covering medium to large-scale models.

\paragraph{Tasks and Datasets.}
To evaluate generalization across task formats, we include both classification and generation tasks. For RoBERTa-large, we follow few-shot classification with $k = 16$ and many-shot classification with $k = 512$ samples per class. We evaluate on 1,000 test examples. For generative models, we use datasets with a consistent 1,000/500/1,000 split for train/validation/test. 

\paragraph{Baselines.}
We compare KerZOO against the state-of-the-art ZO optimization baselines: \textbf{MeZO}~\cite{malladi2023fine}: A memory-efficient ZO method based on symmetric perturbation gradient estimation.
\textbf{HiZOO}~\cite{zhao2024second}: A recent ZO method incorporating approximate second-order curvature via diagonal Hessian estimation.

\paragraph{Implementation Details.}
All experiments are conducted on NVIDIA A100 or A6000 GPUs. For KerZOO, we set the number of perturbation directions to $n = 3$. Hyperparameters such as learning rate and batch size are in line with MeZO baseline. All reported results reflect the best configuration on the validation set.

\begin{table}[htbp]
\centering
\small
\renewcommand{\arraystretch}{1.1}
\captionsetup{justification=raggedright,singlelinecheck=false}
\caption{Performance of various gradient-based and gradient-free optimization methods across multiple datasets and $k$ settings on RoBERTa-large. Bold highlights best performance}
\vspace{0.1in}
\setlength{\tabcolsep}{4pt}
\resizebox{0.95\linewidth}{!}{
\begin{tabular}{llcccccc}
\toprule
\textbf{Task Type} & \textbf{Dataset} & \textbf{SST-2} & \textbf{SST-5} & \textbf{SNLI} & \textbf{MNLI} & \textbf{RTE} & \textbf{TREC} \\
\midrule
\multicolumn{2}{l}{Zero-shot} & 79.0 & 35.5 & 50.2 & 48.8 & 51.4 & 32.0 \\
\cmidrule(lr){1-8}
\multicolumn{8}{c}{\textbf{Gradient-free methods: $k=16$}} \\
\cmidrule(lr){1-8}
MeZO         &        & 90.5 (1.2) & 45.5 (2.0) & 66.0 (2.7) & 56.5 (2.5) & 59.4 (5.3) & 76.9 (2.7) \\
MeZO LoRA    &        & 87.5 (0.7) & 41.6 (0.8) & 64.9 (0.8) & 59.5 (1.5) & 61.7 (3.2) & 58.2 (5.6) \\
KerZOO         &        & \textbf{92.1} (0.8) & \textbf{48.5} (1.0) & \textbf{71.0} (2.2) & \textbf{63.8} (1.8) & \textbf{66.8} (3.3) & \textbf{78.2} (2.4) \\
KerZOO LoRA   &        & 88.4 (1.0) & 42.3 (1.3) & 66.7 (1.7) & 61.1 (1.2) & 63.2 (2.8) & 59.2 (4.9) \\
\cmidrule(lr){1-8}
\multicolumn{8}{c}{\textbf{Gradient-based methods: $k=16$}} \\
\cmidrule(lr){1-8}
FT           &        & 91.9 (1.8) & 47.5 (1.9) & 77.5 (2.6) & 70.2 (2.3) & 66.4 (7.2) & 85.0 (2.5) \\
FT LoRA      &        & 91.4 (1.7) & 46.7 (1.1) & 74.9 (4.3) & 67.7 (1.4) & 66.1 (3.5) & 86.1 (3.3) \\
\cmidrule(lr){1-8}
\multicolumn{8}{c}{\textbf{Gradient-free methods: $k=512$}} \\
\cmidrule(lr){1-8}
MeZO         &        & 93.3 (0.7) & 52.4 (1.2) & 83.0 (1.0) & 78.3 (0.5) & 78.6 (2.0) & 94.3 (1.3) \\
MeZO LoRA    &        & 91.6 (0.8) & 44.8 (0.4) & 73.3 (0.6) & 66.4 (0.4) & 73.3 (1.5) & 63.8 (2.3) \\
KerZOO         &        & \textbf{95.3} (0.5) & \textbf{53.4} (1.0) & \textbf{85.0} (1.5) & \textbf{78.3} (0.5) & \textbf{79.1} (1.8) & \textbf{96.0} (1.9) \\
KerZOO LoRA   &        & 91.9 (0.3) & 45.0 (0.8) & 74.7 (1.1) & 65.0 (0.7) & 74.0 (2.2) & 61.0 (3.2) \\
\cmidrule(lr){1-8}
\multicolumn{8}{c}{\textbf{Gradient-based methods: $k=512$}} \\
\cmidrule(lr){1-8}
FT           &        & 93.9 (0.7) & 55.9 (0.9) & 88.7 (0.8) & 84.4 (0.8) & 82.7 (1.4) & 97.3 (0.2) \\
FT LoRA      &        & 94.2 (0.2) & 55.7 (0.8) & 88.3 (0.5) & 86.9 (0.6) & 83.2 (1.3) & 97.0 (0.3) \\
\bottomrule
\end{tabular}}
\vspace{-0.2in}
\label{tab:main-results}
\end{table}

\begin{table}[htbp]
\centering
\vspace{-0.0in}
\setlength{\tabcolsep}{4pt}
\renewcommand{\arraystretch}{1.1}
\captionsetup{justification=raggedright,singlelinecheck=false}
\caption{Results of fine-tuning OPT-2.7B on seven classification datasets and two generation datasets}
\resizebox{.9\linewidth}{!}{
\begin{tabular}{lcccccccccc}
\toprule
\textbf{Dataset} & \textbf{SST-2} & \textbf{RTE} & \textbf{CB} & \textbf{BoolQ} & \textbf{WSC} & \textbf{WIC} & \textbf{MultiRC} & \textbf{SQuAD} & \textbf{DROP} \\
\textbf{Task Type} & \multicolumn{7}{c}{\emph{classification}} & \multicolumn{2}{c}{\emph{generation}} \\
\midrule
Zero-shot & 56.3 & 54.2 & 50.0 & 47.6 & 36.5 & 52.7 & 44.4 & 29.8 & 10.0 \\
FT        & 94.2 & 81.2 & 82.1 & 72.2 & 63.8 & 65.8 & 71.6 & 78.4 & 30.3 \\
LoRA      & 94.6 & 80.8 & 82.7 & 77.7 & 59.8 & 64.0 & 72.8 & 77.9 & 31.1 \\
\midrule
MeZO      & 91.6 & 63.5 & 69.6 & 67.4 & 62.5 & 59.8 & 59.4 & 63.6 & 15.3 \\
HiZOO     & 90.8 & 60.6 & 70.4 & \textbf{68.0} & 60.2 & 56.6 & 54.8 & 66.0 & \textbf{18.4} \\
\rowcolor{blue!7}
KerZOO     & \textbf{92.6} & \textbf{65.3} & \textbf{71.4} & 67.0 & \textbf{65.4} & \textbf{60.2} & \textbf{62.0} & \textbf{66.2} & 16.0 \\
\midrule
MeZO LoRA     & 91.0 & 63.2 & 69.6 & 67.2 & 64.4 & 58.2 & 59.6 & 57.4 & 13.4 \\
HiZOO LoRA    & 90.6 & \textbf{65.2} & 71.4 & 67.4 & 52.6 & 58.8 & 59.0 & 61.6 & 13.9 \\
\rowcolor{blue!7}
KerZOO LoRA     & \textbf{92.4} & 63.9 & \textbf{73.2} & \textbf{67.4} & \textbf{65.4} & \textbf{60.4} & \textbf{62.4} & \textbf{65.2} &  \textbf{14.7}\\
\bottomrule
\end{tabular}}
\vspace{-0.1in}
\label{tab:opt-2.7}
\end{table}

\subsection{Results on Medium-sized Model}
We assess the performance of KerZOO on RoBERTa-large across multiple classification benchmarks, including SST-2, MNLI, and RTE. We compare against existing zeroth-order (ZO) methods and explore both full-model and parameter-efficient (LoRA-based) tuning.

\begin{figure}[ht]
    \centering
    \begin{subfigure}[t]{0.32\textwidth}
        \centering
        \includegraphics[width=\linewidth]{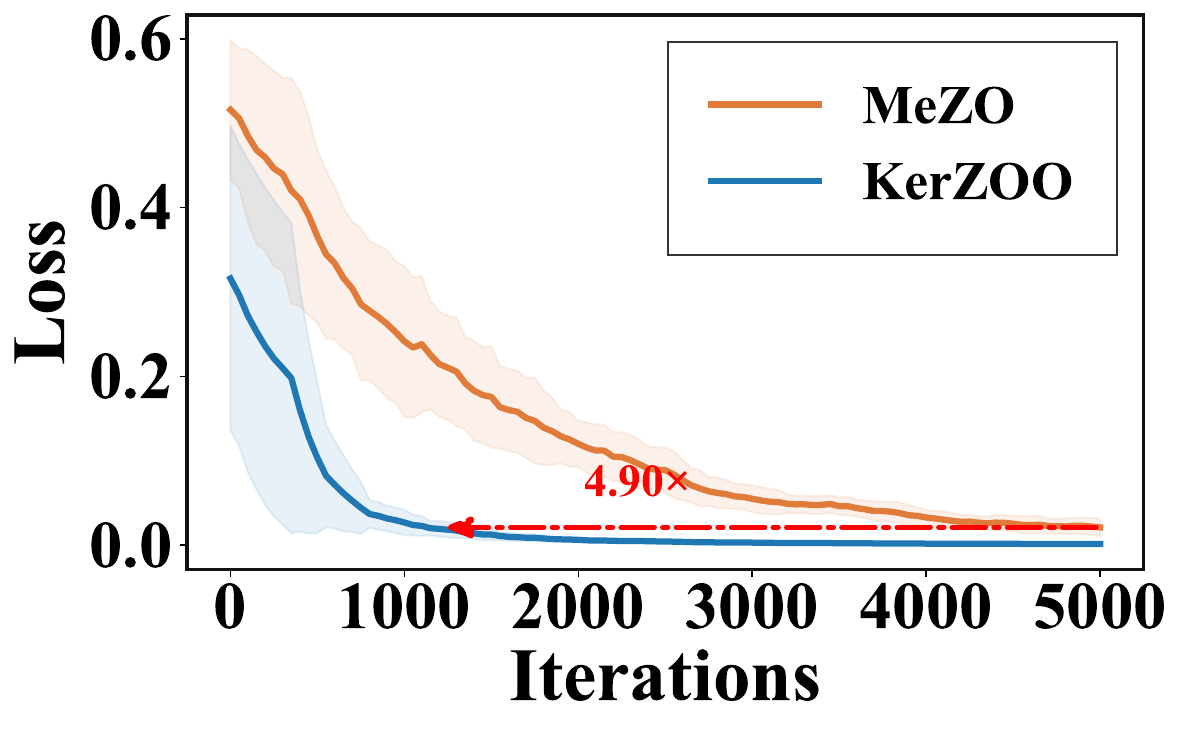}
        \caption{SST-2}
    \end{subfigure}
    \hfill
    \begin{subfigure}[t]{0.325\textwidth}
        \centering
        \includegraphics[width=\linewidth]{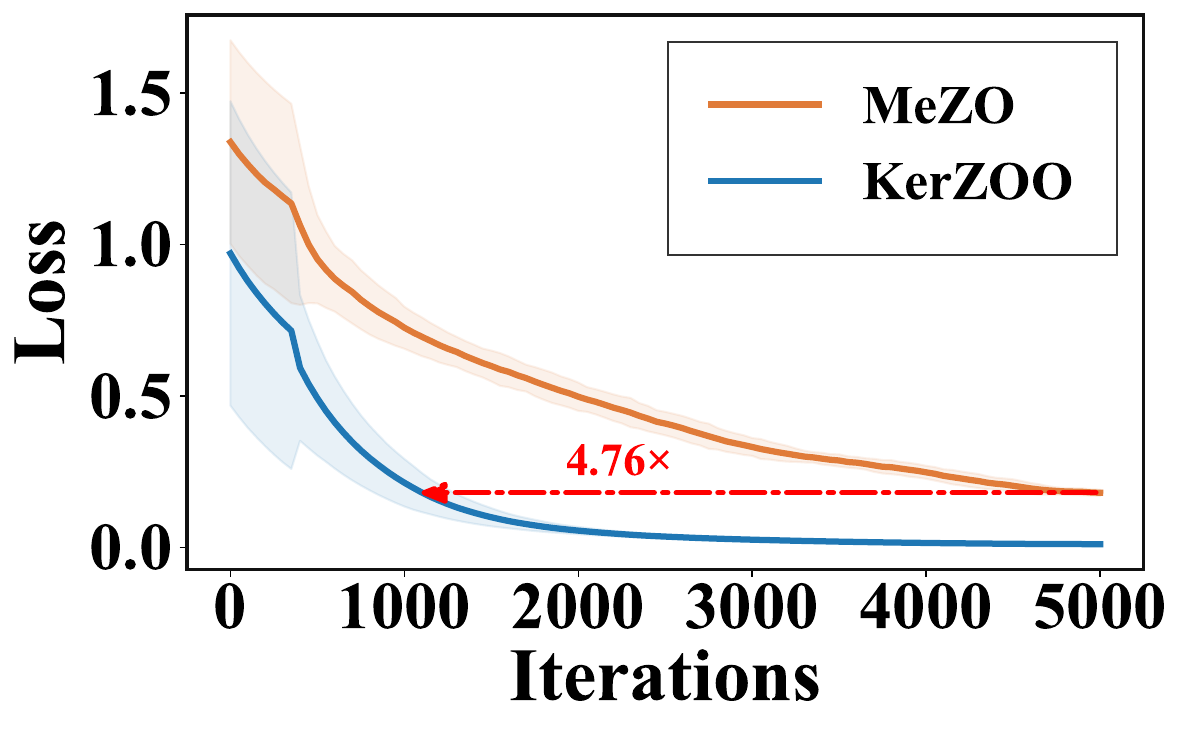}
        \caption{MNLI}
    \end{subfigure}
    \hfill
    \begin{subfigure}[t]{0.32\textwidth}
        \centering
        \includegraphics[width=\linewidth]{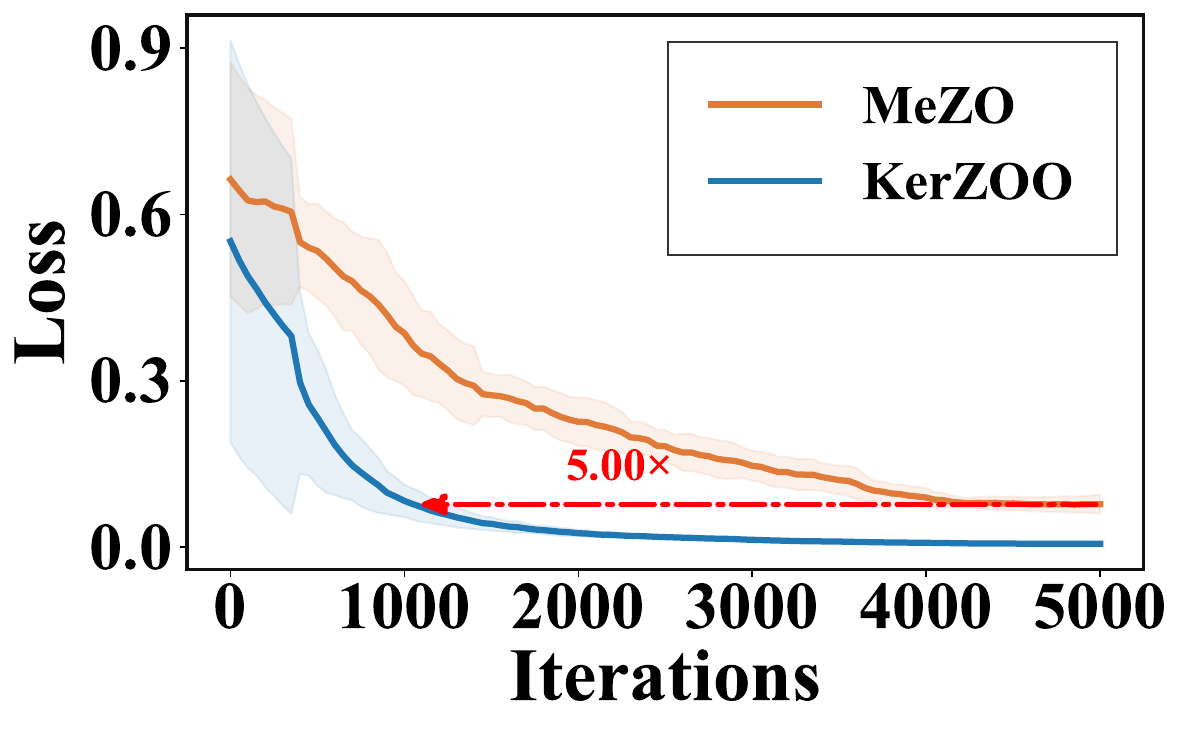}
        \caption{RTE}
    \end{subfigure}
    \caption{Training loss comparison of MeZO and KerZOO (Ours) on RoBERTa-large}
    \label{fig:rob_loss}
\end{figure}

\noindent\textbf{Faster Optimization.}
KerZOO achieves a markedly faster descent in training loss compared to MeZO, as shown in Figure~\ref{fig:rob_loss}. When using only three perturbation directions, our method reduces training iterations by over 70\% on average, and lowers wall-clock convergence time by 30\%--40\% on SST-2, MNLI, and RTE. This acceleration stems from the improved stability of our kernel-based gradient estimation.

\noindent\textbf{Improved Accuracy.}
In addition to faster convergence, KerZOO also delivers consistently higher task accuracy. On the three datasets mentioned above, it improves upon MeZO by 1.7\%, 7.3\%, and 7.4\%, respectively. The results are summarized in Table~\ref{tab:main-results}. In several cases—such as RTE and SST-5—KerZOO even matches or exceeds the accuracy of first-order fine-tuning.
We further evaluate KerZOO under the LoRA framework to test its compatibility with parameter-efficient tuning. While LoRA generally introduces a small performance drop compared to full fine-tuning, KerZOO remains competitive and continues to outperform ZO baselines in most cases. This highlights KerZOO’s robustness under limited trainable parameter budgets.

\begin{figure}
    \centering
    \includegraphics[width=0.99\textwidth]{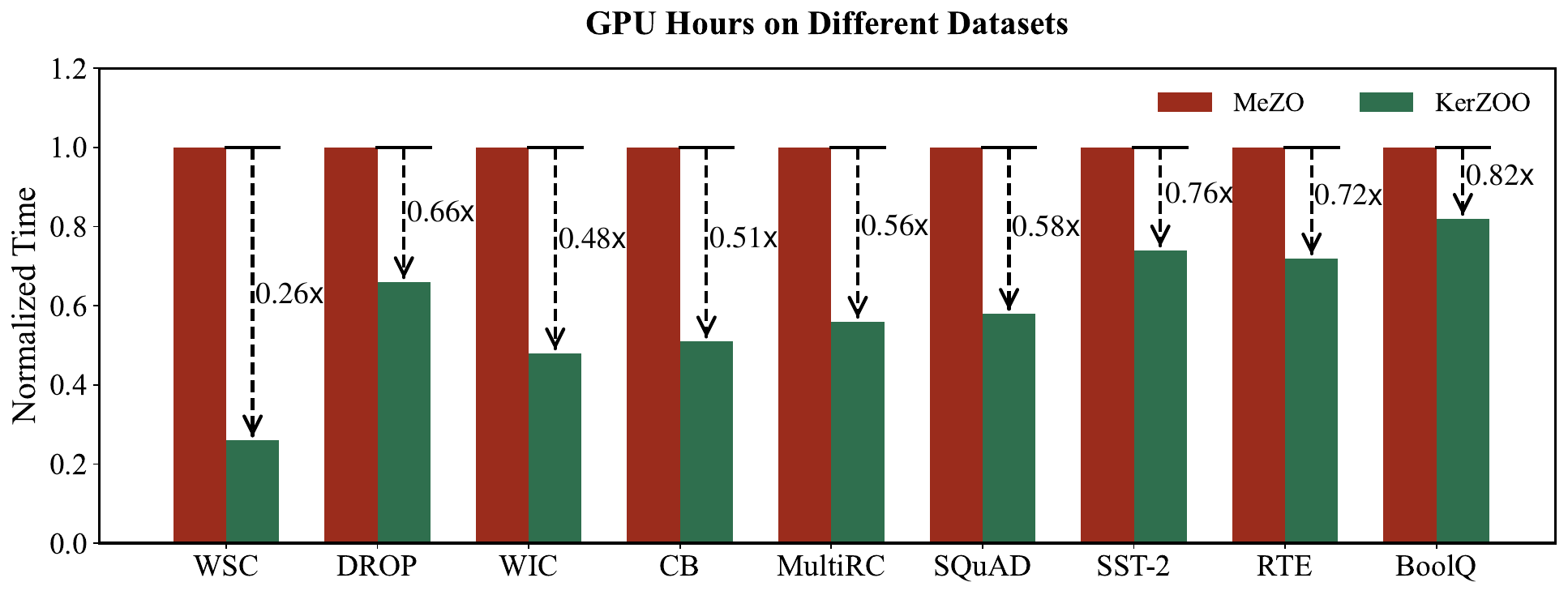}
    \captionsetup{justification=raggedright, singlelinecheck=false}
    \vspace{-0.1in}
    \caption{Comparison of GPU hours for convergence across different datasets on OPT-2.7B between MeZO and KerZOO. Results are presented as normalized time}
    \vspace{-0.2in}
    \label{fig:time}
\end{figure}

\subsection{Results on Autoregressive Models (OPT \\and LLaMa)}

\begin{wrapfigure}[4]{r}{0.45\textwidth}
\vspace{-1in}
\centering
\includegraphics[width=0.95\linewidth]{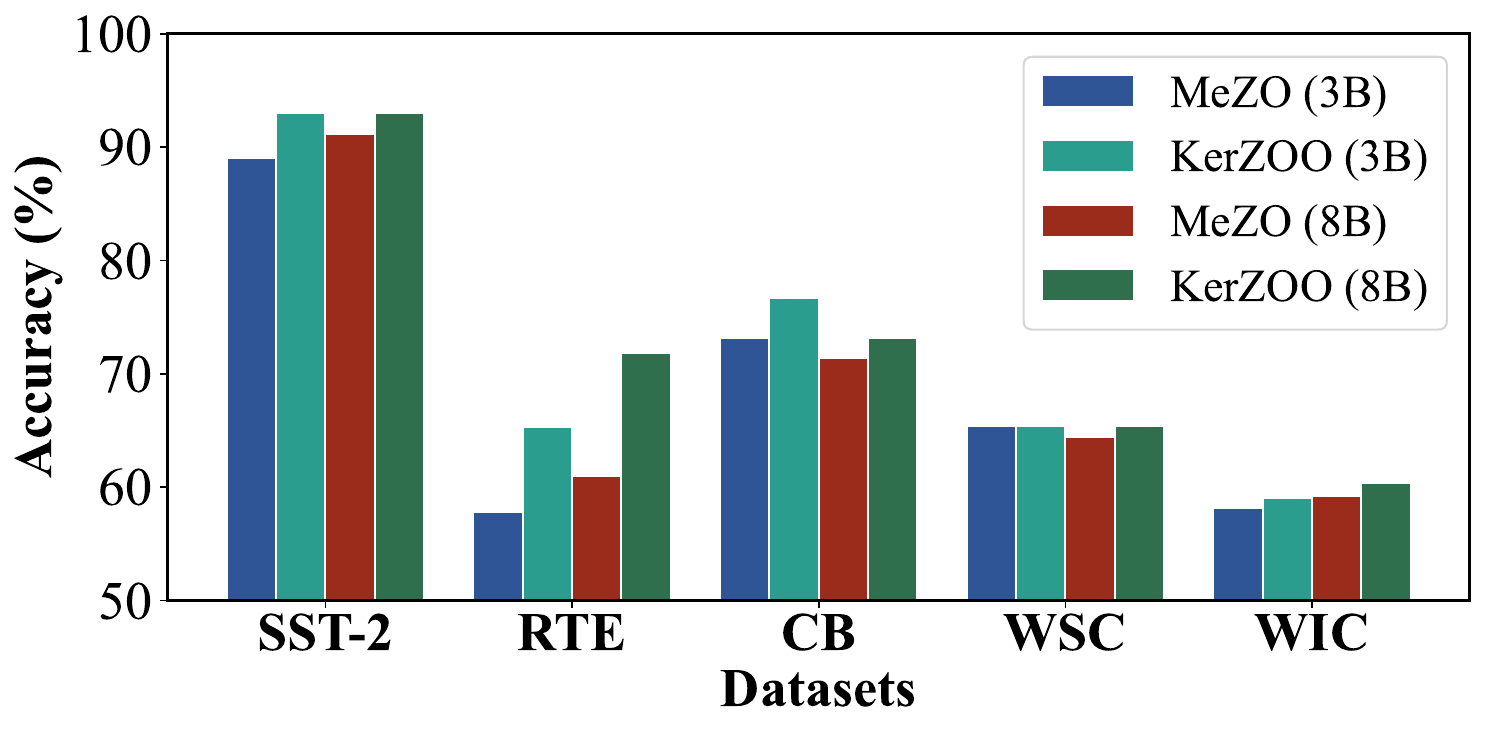}
\vspace{-0.1in}
\captionsetup{justification=raggedright,singlelinecheck=false}
\caption{Results on LLaMA3-3B and LLaMA3-8B}
\label{fig:llama}
\vspace{-0.1in}
\end{wrapfigure}

We evaluate KerZOO on OPT and LLaMA series models across classification and generation tasks. The experiments span OPT-2.7B, OPT-6.7B, and LLaMA3-3B/8B models. Results are reported in Table~\ref{tab:opt-2.7}, Table~\ref{tab:opt-6.7}, and Figure~\ref{fig:llama}, respectively. We also compare GPU time efficiency and memory usage on the SQuAD dataset in Table~\ref{tab:gpu_hour}.

Across all model scales, KerZOO consistently reduces the training time required to reach competitive performance. As shown in Figure~\ref{fig:time}, it completes training with notably fewer update steps compared to MeZO—achieving up to a 74\% reduction in computing time without reliance on gradient backpropagation.
KerZOO offers consistent performance gains under both full-model fine-tuning and LoRA-based adaptation. On OPT-2.7B, it outperforms zeroth-order baselines in 6 of 7 classification tasks, and remains competitive on generation benchmarks. Table~\ref{tab:opt-6.7} shows that these trends persist when moving to larger models such as OPT-6.7B. When applied to LLaMA-3 variants, KerZOO also yields robust improvements, including about 10\% accuracy gains on RTE (Figure~\ref{fig:llama}).

\begin{wraptable} [11]{r}{.49\linewidth}  
\vspace{-0.15in}
\centering
\setlength{\tabcolsep}{2pt}
\renewcommand{\arraystretch}{0.9}
\captionsetup{justification=raggedright,singlelinecheck=false}
\caption{Experiment results on OPT-6.7B for four classification datasets and one text generation dataset (with 1000 training samples)}
\resizebox{\linewidth}{!}{
\begin{tabular}{lccccc}
\toprule
\textbf{Dataset} & \textbf{SST-2} & \textbf{RTE} & \textbf{CB}  & \textbf{WSC}  & \textbf{SQuAD}  \\
\textbf{Task Type} & \multicolumn{4}{c}{\emph{classification}} & \multicolumn{1}{c}{\emph{generation}} \\

\midrule
MeZO      & 92.2 & 69.6 & 73.2 & 60.6 &  68.9 \\
HiZOO     & 90.8 & 66.3 & 71.8 &  62.1 & \textbf{71.9} \\
\rowcolor{blue!7}
KerZOO     & \textbf{94.0} & \textbf{71.9} & \textbf{78.6} & \textbf{65.4} & 70.7  \\
\midrule
MeZO LoRA     & 93.2 & 67.9 & \textbf{73.2} & 59.6 & 63.4 \\
HiZOO LoRA    & 90.6 & 67.2 & 71.4 & 62.1 &  65.3 \\
\rowcolor{blue!7}
KerZOO LoRA     & \textbf{93.8} & \textbf{68.6} & \textbf{73.2} & \textbf{62.5} & \textbf{72.1} \\
\bottomrule
\end{tabular}}

\label{tab:opt-6.7}
\end{wraptable}

Moreover, KerZOO maintains memory efficiency compared to full fine-tuning. As detailed in Table~\ref{tab:gpu_hour}, its memory footprint increases marginally but can achieve significant decrease of traning GPU hours. Compare to MeZO and HiZOO, our training time can decrease about 42\% and 33\% respectively in full fine-tuning on SQuAD datasets. In PEFT settings, KerZOO paired with LoRA achieves the best balance between memory cost and convergence speed — consuming 36\% of the GPU hours required by MeZO while yielding higher accuracy.

\vspace{-0.8em}
\section{Analysis on Memory and Speed}
\vspace{-0.2em}

\begin{wraptable}[14]{r}{0.49\textwidth}  
\vspace{-0.15in}  
\centering
\renewcommand{\arraystretch}{1.1}
\captionsetup{justification=raggedright,singlelinecheck=false}
\caption{Memory and training time comparison on OPT-2.7B (SQuAD, avg. 300 tokens)}
\label{tab:gpu_hour}

\resizebox{0.45\textwidth}{!}{
\begin{tabular}{lccc}
\toprule
\textbf{Method} & \textbf{Mem.} & \textbf{Iter.} & \textbf{Hours} \\
\midrule
FT         & 73.5G & 7.5\%   & 27.7\% \\
LoRA       & 58.5G & 6.3\%   & 11.5\% \\
\midrule
MeZO       & 12.8G & 100.0\% & 100.0\% \\
HiZOO      & 14.1G & 66.7\%  & 91.5\% \\
KerZOO     & 16.9G & 16.9\%  & 58.0\% \\
\midrule
MeZO+LoRA  & 8.1G  & 94.2\%  & 51.6\% \\
HiZOO+LoRA & 9.1G  & 80.0\%  & 65.7\% \\
KerZOO+LoRA& 9.7G  & 28.4\%  & 36.1\% \\
\bottomrule
\end{tabular}
}
\vspace{-1.0em}
\end{wraptable}

We evaluate KerZOO's memory cost and training efficiency under both full-parameter and LoRA-based tuning regimes. As shown in Table~\ref{tab:gpu_hour}, KerZOO achieves a favorable trade-off between convergence speed and per-step overhead. While its per-step memory cost (16.9G) is slightly higher than MeZO (12.8G) due to the use of multiple perturbations, it substantially reduces the number of required iterations from 100\% to only 16.9\%, resulting in a total GPU hour reduction from 100.0\% to 58.0\%. On the SQuAD dataset, for instance, KerZOO lowers training time by over 40\% compared to MeZO while maintaining competitive accuracy.

From a memory perspective, KerZOO avoids storing gradients and activations, making it significantly more efficient than gradient-based methods such as full fine-tuning (73.5G) and LoRA (58.5G). 
Even under parameter-efficient settings, KerZOO maintains strong performance: KerZOO+LoRA reduces GPU hours to 36.1\%, compared to 51.6\% with MeZO+LoRA and 65.7\% with HiZOO+LoRA, while using 9.7G memory. These results highlight that KerZOO offers efficient convergence with minimal memory overhead across a wide range of configurations.

\section{Related Work}

\subsection{Memory-efficient LLM Fine-tuning}
\vspace{-0.1in}

Fine-tuning pre-trained models~\cite{devlin2018bert,liu2019roberta,chen2022visualgpt,radford2021learning,singh2022revisiting} has emerged as an effective strategy for leveraging previously acquired representations.
However, it remains resource-intensive especially with the scaling model size, prompting the development of more memory-efficient alternatives. Parameter-efficient fine-tuning (PEFT) techniques, including LoRA~\cite{hu2021lora} and prefix tuning~\cite{li2021prefix}, address this challenge by modifying only a limited portion of model parameters
, thereby retaining most of the original pre-trained weights and the knowledge they encode. 
Low-rank adaptation methods which exemplified by LoRA, have shown strong performance by introducing low-rank updates that are lightweight yet effective. Building on this, DoRA~\cite{liu2024dora} further decomposes pre-trained weights into direction and magnitude components to improve both expressiveness and training stability. GaLore~\cite{zhao2024galore} introduces gradient low-rank projection to enable full-parameter training while maintaining the memory efficiency characteristic of low-rank updates.
In addition to low-rank techniques, quantization has become a key strategy for reducing resource usage. 
SmoothQuant~\cite{xiao2023smoothquant} further optimizes mixed-precision training by smoothing out activation outliers through offline scaling, shifting quantization difficulty from activations to weights. Outlier Suppression+~\cite{wei2023outlier} enhances low-bit (e.g., int4) quantization by applying channel-wise shifting and scaling to reduce asymmetry and variance in activation distributions.

\subsection{Zeroth-order Optimization and Acceleration}
\vspace{-0.1in}

Zeroth-order optimization, a classical optimization schema that uses only differences of loss values for gradient estimation, has attracted significant attention in the machine learning community~\cite{chen2017zoo, ye2018hessian, verma2023certified, chen2023deepzero, malladi2023fine}. Unlike conventional first-order optimization, which computes the gradient via cumbersome backpropagation, the ZO method can, in principle, update the model with just forward passes.

Recently, the ZO method has been proven to be efficient in solving the significant memory limitation in large-scale LLMs fine-tuning~\citep{malladi2023fine,liu2024sparse,tang2024private,zhao2024second}. 
However, ZO optimization often converges more slowly than FO approaches, primarily due to the noise from its randomized estimators. A variety of strategies have been proposed to improve the efficiency of ZO optimization. \citep{liu2018zeroth} introduces ZO-SVRG by integrating variance reduction methods~\cite{johnson2013accelerating}, helping to stabilize gradient estimates. \citep{shu2023zeroth} employs Gaussian process surrogates to approximate the objective landscape, thereby reducing query overhead and enabling denser sampling. \citep{li2024pretrained} advances the idea of pretrained optimizers, which transfer knowledge across tasks to enable rapid zero-shot adaptation within a few ZO fine-tuning steps. In parallel, \citep{zhao2024second} proposes HiZOO, a framework that incorporates second-order information via estimated Hessians to guide more effective updates. Despite these efforts, adapting ZO methods to LLMs still faces challenges. Many accelerator were designed for small-scale tasks and lack scalability. Moreover, as recent work suggests single-step tuning can suffice~\cite{malladi2023fine}, the focus needs to shifts from query efficiency to stability. Additionally, some accelerators also compromise ZO’s key benefits in low memory usage and high throughput. These challenges underscore the need for scalable, efficient ZO strategies for LLM fine-tuning.


\section{Limitations} \label{sec:limitations}

\vspace{-0.1in}
While our current experiments primarily focus on LLM fine-tuning, we envision future extensions of zeroth-order optimization techniques to a wider range of AI domains, such as vision-language model adaptation, and the pruning or quantization of large-scale language models.

\section{Conclusion} \label{conc}

\vspace{-0.1in}

In this work, we present a theoretical framework that characterizes the lower-order bias arising in gradient estimation when using zeroth-order optimization methods. Building on this analysis, we propose a Kernel Function Informed Zeroth-Order Optimization method, an improved ZO framework that leverages kernel function to eliminate the lower-order bias of zeroth-order gradient estimation. KerZOO offers notable improvements in training efficiency and consistently outperforms baseline approaches across a variety of tasks and different models. Meanwhile, KerZOO is fully compatible with parameter-efficient tuning (PEFT) techniques such as LoRA, allowing for additional acceleration without degrading accuracy. Looking ahead, we plan to extend this framework to other domains, with a particular focus on adapting it for LLM pruning or vision language model optimization.

\bibliographystyle{unsrt}
\bibliography{neurips_2025}            

\newpage

\section{Appendix} \label{app}

\subsection{Variance Analysis of the Zeroth-Order Estimator with Kernel Function} \label{var}

When applying a kernel function $K(r)$ and a scalar random variable $r$ in the zeroth-order estimator, the gradient estimate can often be written in the form:
\begin{equation}
\hat{g_K} \approx \langle \nabla \mathcal{L}(\bm{\mathcal{\theta}}), \bm{u} \rangle \cdot \bm{u} \cdot r K(r) \approx M \cdot rK(r),
\end{equation}
where $M$ approximates a constant related to the gradient magnitude and perturbation step size and $\bm{u}$ is a unit vector sampled uniformly at random.
Under the formulation, we main focus is how the product $rK(r)$ affects the variance of the estimator. We know that variance is defined as:
\begin{equation}
\mathrm{Var}(X) = \mathbb{E}[X^2] - (\mathbb{E}[X])^2.
\end{equation}
If the estimator is approximately unbiased, i.e., $\mathbb{E}[\hat{g_K}] \approx c\nabla f(x)$ (c is a constant), 
we focus on
$\mathbb{E}[\hat{g_K}^2]$. 

In our estimator, the randomness comes from the term $r K(r)$. From the derivation above, we can rewrite the estimator in a simplified form:
\begin{equation}
\hat{g} = M \cdot (r K(r)).
\end{equation}
Then its second moment becomes:
\begin{equation}
\mathbb{E}[\hat{g_K}^2] \propto \mathbb{E}\left[(r K(r))^2\right].
\end{equation}

Since $\bm{u}$ is a unit vector (or normalized), the fluctuation in the estimator is primarily greatly related to the term $rK(r)$. If $\mathbb{E}[(rK(r))^2]$ is large, then even with fixed $M$ and $\bm{u}$, the overall variance of the estimator will be large. Conversely, a smaller value of $\mathbb{E}[(rK(r))^2]$ helps reduce the estimator's variance. Since we use only three random perturbations to estimate the gradient, the variance can be relatively large in such low-perturbation settings. To address this, we constrain $r$ within the interval $[-1, 1]$, and gradually shrink the range of $r$ as the number of iterations increases. This strategy enables a trade-off between the variance and bias of the gradient estimation (Restricting the range of 
$r$ may result in gradient estimates that are not strictly unbiased; however, it effectively reduces the variance of the estimate when the number of perturbations is limited.).

\subsection{Complementary experimental settings}

\textbf{Baselines.} We select two representative baseline methods for our evaluation, i.e., MeZO~\cite{malladi2023fine} and HiZOO~\cite{zhao2024second}. MeZO~\cite{malladi2023fine} is a memory-efficient approach for LLM fine-tuning that avoids storing full perturbation vectors by regenerating them using a fixed seed. This eliminates the need for extra memory allocation and simplifies implementation. 
HiZOO~\cite{zhao2024second} enhances convergence efficiency by incorporating approximate second-order curvature information into the optimization process, thereby addressing the typically slow convergence characteristic of first-order ZO methods.

\textbf{Hyperparameter settings.} Our experiments on RoBERTa-large, the OPT family, and LLaMA models adopt the hyperparameter configurations listed in Table \ref{tab:hyperparams}. We also keep the hyperparameters consistent with MeZO, such as the learning rate and the perturbation scale. In our experiments, our kernel function constant $C$ is set to 4.

\begin{table}[htbp]
\centering
\caption{The hyperparameters setting in our experiments. 
}
\vspace{0.05in}
\resizebox{\textwidth}{!}{
\begin{tabular}{lll}
\toprule
\textbf{Experiment} & \textbf{Hyperparameters} & \textbf{Values} \\
\midrule
\multirow{3}{*}{FT} 
    & Batch size         & 8 \\
    & Learning rate      & \{1e-5, 5e-5\} \\
    & Lr schedule        & Constant for RoBERTa; Linear for OPT and LLaMA \\
\midrule
\multirow{4}{*}{MeZO}
    & Batch size         & \{64, 16\} \\
    & Learning rate $\eta$ (Lr) & \{1e-6, 5e-7\} \\
    & $\epsilon$         & 1e-3 \\
    & Lr schedule        & Constant for RoBERTa; Linear for OPT and LLaMA \\
\midrule
\multirow{4}{*}{MeZO LoRA}
    & Batch size         & \{64, 16\} \\
    & Learning rate $\eta$ (Lr) & \{1e-4, 5e-5\} \\
    & $\epsilon$         & 1e-2 \\
    & Lr schedule        & Constant for RoBERTa; Linear for OPT and LLaMA \\
\midrule
\multirow{3}{*}{KerZOO (LoRA)}
    & Kernel function order $\beta$ & 3 \\
    & $r$ & Shrink as iteration step increases \\
    & Kernel function constant $C$        & 4 \\
\bottomrule
\end{tabular}
}
\label{tab:hyperparams}
\end{table}

\textbf{Datasets.} In the RoBERTa-large experiments, we employ a range of classification benchmarks, including SST-2, SST-5, SNLI, TREC, MNLI, and RTE. These benchmarks cover a range of sentence-level and textual entailment tasks from previous NLP studies~\cite{socher2013recursive, bowman2015large, voorhees2000building, yao2020improving, dagan2005pascal, barhaim2006second, bentivogli2009fifth, giampiccolo2007third}. To maintain consistency with prior works~\cite{malladi2023fine,zhao2024second}, we set the test set size as 1000 examples. We examine both few-shot and many-shot regimes, setting the number of training examples per class to $k=16$ or $k=512$ and using the same number for validation. For experiments involving the OPT and LLaMA model families, we use the SuperGLUE benchmark~\cite{wang2019superglue}, which includes tasks such as RTE~\cite{dagan2005pascal, barhaim2006second, bentivogli2009fifth, giampiccolo2007third}, CB~\cite{demarneffe2019shared}, BoolQ~\cite{clark2019boolq}, WIC~\cite{pilehvar2018wic}, WSC~\cite{levesque2012winograd}, and MultiRC~\cite{khashabi2018looking}. In addition, we include SST-2~\cite{socher2013recursive} and two QA datasets, SQuAD~\cite{rajpurkar2016squad} and DROP~\cite{dua2019drop}. For each dataset, we randomly select 1000 samples for training, 500 for validation, and 1000 for evaluation.

\subsection{More results}

\subsubsection{More Results on LLaMa-3 model}

We conduct fine-tuning experiments of KerZOO on the LLaMA-3 model series. We use exactly the same hyperparameter settings as those used for the OPT family of models. The detailed results of the experiments are shown in Table \ref{tab:llama3-3b} and \ref{tab:llama3-8b} below. 

\begin{table}[htbp]
\centering
\caption{Experiment results on LLaMA3-3B (1000 training samples)}
\vspace{0.5em}

\begin{tabular}{lccccc}
\toprule
\textbf{Task} & \textbf{SST-2} & \textbf{RTE} & \textbf{CB} & \textbf{WSC} & \textbf{WIC} \\
\midrule
FT     & 94.2 & 81.2 & 91.4 & 72.2 & 63.8 \\
MeZO   & 89.0 & 57.8 & 73.2 & \textbf{65.4} & 58.2 \\
\rowcolor{blue!7}
KerZOO   & \textbf{93.0} & \textbf{65.3} & \textbf{76.7} & \textbf{65.4} & \textbf{59.0} \\
\bottomrule
\end{tabular}

\label{tab:llama3-3b}
\end{table}

\begin{table}[htbp]
\centering
\caption{Experiment results on LLaMA3-8B (1000 training samples)}
\vspace{0.5em}

\begin{tabular}{lccccc}
\toprule
\textbf{Task} & \textbf{SST-2} & \textbf{RTE} & \textbf{CB} & \textbf{WSC} & \textbf{WIC} \\
\midrule
MeZO   & 91.2 & 61.0 & 71.4 & 64.4 & 59.2 \\
\rowcolor{blue!7}
KerZOO   & \textbf{93.0} & \textbf{71.8} & \textbf{73.2} & \textbf{65.4} & \textbf{60.4} \\
\bottomrule
\end{tabular}

\label{tab:llama3-8b}
\end{table}

\subsubsection{More Analysis on Memory and Speed }

\begin{table}[htbp]
\centering
\renewcommand{\arraystretch}{1.1}
\captionsetup{justification=raggedright, singlelinecheck=false}
\caption{Memory and training time comparison of OPT-2.7B on SST-2 dataset (35 tokens per example on average)}

\vspace{0.05in}
\label{tab:gpusst}

\begin{tabular}{lccc}
\toprule
\textbf{Method} & \textbf{Memory cost} & \textbf{Iteration step} & \textbf{GPU hours} \\
\midrule
FT         & 45.4G & 9.3\%   & 16.8\% \\
LoRA       & 18.5G & 5.6\%   & 4.3\% \\
\midrule
MeZO       & 10.7G & 100.0\% & 100.0\% \\
HiZOO      & 11.3G & 59.2\%  & 87.4\% \\
KerZOO     & 14.7G & 25.0\%  & 76.2\% \\
\midrule
MeZO+LoRA  & 5.5G  & 74.1\%  & 43.7\% \\
HiZOO+LoRA & 5.7G  & 46.3\%  & 41.0\% \\
KerZOO+LoRA& 5.7G  & 16.3\%  & 29.5\% \\
\bottomrule
\end{tabular}
\end{table}

\begin{table}[htbp]
\centering
\renewcommand{\arraystretch}{1.1}
\captionsetup{justification=raggedright, singlelinecheck=false}
\caption{Memory and training time comparison of OPT-2.7B on RTE dataset (180 tokens per example on average)}

\vspace{0.05in}
\label{tab:gpurte}

\begin{tabular}{lccc}
\toprule
\textbf{Method} & \textbf{Memory cost} & \textbf{Iteration step} & \textbf{GPU hours} \\
\midrule
FT         & 62.2G & 10.0\%   & 16.2\% \\
LoRA       & 42.5G & 8.3\%   & 6.6\% \\
\midrule
MeZO       & 13.5G & 100.0\% & 100.0\% \\
HiZOO      & 13.8G & 63.3\%  & 88.9\% \\
KerZOO     & 14.5G & 22.5\%  & 72.3\% \\
\midrule
MeZO+LoRA  & 7.5G  & 73.3\%  & 34.8\% \\
HiZOO+LoRA & 7.8G  & 56.7\%  & 35.9\% \\
KerZOO+LoRA& 7.7G  & 7.6\%  & 11.5\% \\
\bottomrule
\end{tabular}
\end{table}

We conduct experiments on the convergence and memory consumption of KerZOO on the OPT-2.7B model using the SST-2 and RTE datasets. As anticipated in our earlier analysis, KerZOO introduces only a slight increase in memory usage compared to the MeZO and HiZOO baselines. However, it exhibits fast convergence. For instance, on the RTE dataset, KerZOO combined with LoRA requires only about 11\% of the training time required by MeZO under full fine-tuning, while still achieving competitive accuracy.


\begin{figure}
    \centering
    \includegraphics[width=0.8\textwidth]{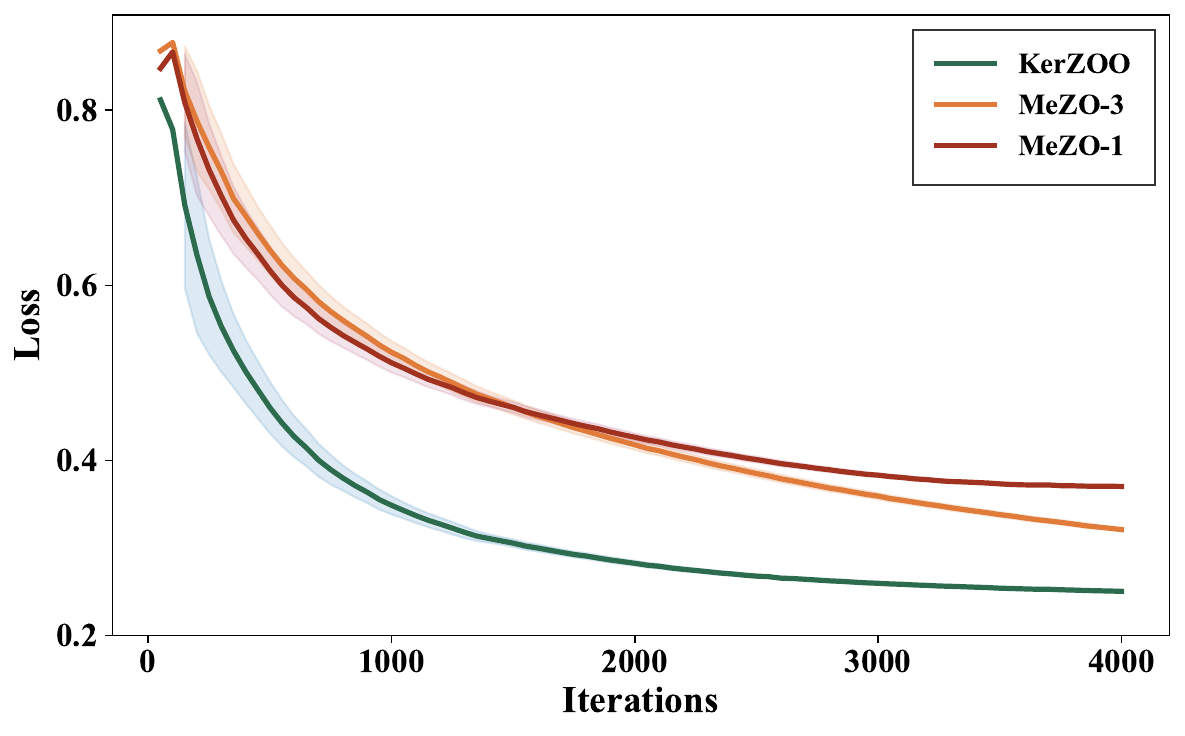}
    \vspace{-0.1in}
    \caption{Training loss curves of MeZO with different perturbations and KerZOO}
    \vspace{-0.2in}
    \label{fig:compare}
\end{figure}

We show the training loss curves of the OPT-2.7B model on the SST-2 dataset in Figure~\ref{fig:compare}. The comparison includes the original MeZO method with one perturbation per update (MeZO-1), MeZO with three perturbations (MeZO-3), and our proposed KerZOO method. Consistent with the findings reported in the original MeZO paper, we can observe that increasing the number of perturbations to three brings only marginal improvement in convergence. 
In contrast, our KerZOO demonstrates significantly faster convergence, 
highlighting the effectiveness of our kernel-based ZO optimization.

\subsection{Convergence analysis}

In this section, we give analysis on the convergence of KerZOO based on the former works on the kernel function \cite{bychkov2024accelerated, lobanov2023accelerated}. We use $\|\bm{\theta}\|_p$ to denote the $\ell_p$-norm of the parameter $\bm{\theta}$ in LLMs, we use the notation $\|\bm{\theta}\|_2 = \|\bm{\theta}\|$ to express the Euclidean norm for simplicity.
$\langle \bm{\theta}, \bm{\eta} \rangle := \sum_{k=1}^d \theta_i \eta_i$ is denoted as the standard inner product of $\bm{\theta}, \bm{\eta} \in \mathbb{R}^d$, where $\theta_i$ and $\eta_i$ are the $i$-th component of $\bm{\theta}$ and $\bm{\eta}$, respectively. $S_p^d(r) := \{\bm{\theta} \in \mathbb{R}^d : \|\bm{\theta}\|_p = r\}$ and $B_p^d(r) := \{\bm{\theta} \in \mathbb{R}^d : \|\bm{\theta}\|_p \leq r\}$ are used to express the sphere and the ball in the $\ell_p$-norm. Notation $\lesssim$ is used to denote the asymptotic inequality.

\paragraph{Assumption 1 ($L$-smoothness).}
Suppose loss function $\mathcal{L}(\bm{\theta})$ is an $L$-Lipschitz smooth function, or it has $L$-Lipschitz continuous gradient. If $\mathcal{L}(\bm{\theta})$ is continuously differentiable with respect to $\bm{\theta}$, and its gradient satisfies the Lipschitz condition for any $\xi$ and $\bm{\theta}, \bm{\eta} \in \mathbb{R}^d$
\begin{equation}
\|\nabla \mathcal{L}(\bm{\theta}) - \nabla \mathcal{L}(\bm{\eta})\| \leq L \|\bm{\theta} - \bm{\eta}\|.
\end{equation}

\paragraph{Assumption 2 (Higher-order smoothness).}
Let $t$ denote the maximal integer which is strictly less than $\beta$. Let $\mathcal{F}_\beta(L)$ denote the set of all functions $\mathcal{L} : \mathbb{R}^d \to \mathbb{R}$ which can be derived for $t$ times and for any $\bm{\theta}, \bm{\eta} \in \mathbb{R}^d$ satisfy the Hölder-type condition:
\begin{equation}
\left| \mathcal{L}(\bm{\theta}) - \sum_{0 \leq |n| \leq t} \frac{1}{n!} D^n \mathcal{L}(\bm{\eta})(\bm{\theta} - \bm{\eta})^n \right| \leq L_\beta \|\bm{\theta} - \bm{\eta}\|,
\end{equation}
where $L_\beta > 0$, the sum is over multi-index $n = (n_1, \ldots, n_d) \in \mathbb{N}^d$, the notation $n! = n_1! \cdots n_d!$, and
\begin{equation}
D^n \mathcal{L}(\bm{\eta}) \bm{v}^n = \frac{\partial^{|n|} \mathcal{L}(\bm{\eta})}{\partial \theta_1^{n_1} \cdots \partial \theta_d^{n_d}} v_1^{n_1} \cdots v_d^{n_d},
\end{equation}
where $|n| = n_1 + \cdots + n_d$, for all $\bm{v} = (v_1, \ldots, v_d) \in \mathbb{R}^d$.

Assumption 1 is a special case of Assumption~2 if $\beta=2$.

Based on\cite{woodworth2021even}, we have the following assumption:  

\paragraph{Assumption 3.} 
There exists $\sigma_*^2 \geq 0$ such that
\begin{equation}
\mathbb{E}\left[\|\mathbf{g}^* - \nabla \mathcal{L}(\bm{\theta}^*)\|^2\right] \leq \sigma_*^2.
\end{equation}

which means that in overparameterized regime the variance of our estimation of stochastic gradient $\mathbf{g}^*$ at optimal point 
$\bm{\theta}^* = \arg\min_{\bm{\theta} \in \mathbb{R}^d} \mathcal{L}(\bm{\theta})$ 
can be upper bounded by using the minimum value $\mathcal{L}(\bm{\theta}^*)$. We will use this assumption for the gradient estimates.

Then by using two-point noisy zero-order oracle (Assumption~2), the gradient can be estimated as follows:
\begin{equation}
\mathbf{g}(\bm{\theta}, \bm{u}, r) = \frac{1}{2\epsilon} \left( \mathcal{L}(\bm{\theta} + \epsilon r \bm{u}) - \mathcal{L}(\bm{\theta} - \epsilon r \bm{u}) \right) K(r) \bm{u},
\end{equation}

where $\epsilon > 0$ is a smoothing parameter (perturbation scale), $\bm{u} \in S_2^d(1)$ is a vector which is defined in Section \ref{method}, 
$K : [-1, 1] \to \mathbb{R}$ is a kernel function that satisfies

\[
\mathbb{E}[K(a)] = 0, \quad 
\mathbb{E}[a K(a)] = C, \quad 
\mathbb{E}[|a| |K(a)|] < \infty, \quad 
\text{and for all } j = 2, \dots, t,\quad \mathbb{E}[a_j K(a)] = 0.
\]

With Assumption 2, we will give a theorem for stating the convergence of KerZOO for LLM with two-point zeroth-order estimation. We denote:

\[
\kappa_\beta = \int |a|^\beta |K(a)| \, da, \quad 
\text{and} \quad 
\kappa = \int |K(a)|^2 \, da.
\]

\paragraph{Theorem.}
Let $\mathcal{L}(\cdot)$ satisfy Assumption~2 with parameter $\beta$. Let smoothing parameter be 
$\epsilon \leq \left(\frac{\psi}{\kappa_\beta L R}\right)^{1/(\beta - 1)}$, then we use our Algorithm 1 to do gradient estimation.
Let $\bm{\theta}_N^{ag}$ be the output of Algorithm 1, then
\begin{equation}
\mathbb{E}\left[\mathcal{L}(\bm{\theta}_N^{ag}) - \mathcal{L}(\bm{\theta}^*)\right] \leq \psi
\end{equation}
in at most $N$ iterations and $T$ oracle calls. Note that in our case, $T$ equals to $N$ in our setting. $\rho$ denotes the radius to the optimal solution $\bm{\theta}^*$ such that 
$\|\bm{\theta}^* - \bm{\theta}_0\| \leq \rho$, and $\bm{\theta}_0$ is a starting point. Then, we have

   \begin{equation}
    N =T= \mathcal{O}\left( \max\left( \frac{L \rho^2}{\psi}, \frac{ \kappa \sigma_*^2 \rho^2}{\psi^2} \right) \right)
    \end{equation}

\paragraph{Analysis.}
Now we should find the upper bounds for smoothing
parameter $\epsilon$, the asymptotic for iteration steps $N$ (number of oracle calls $T$).
We consider to limit the bias and second moment of estimation of gradients. (1) For the bias of gradient approximation, it can be expressed as
\begin{equation}
b = \left\| \mathbb{E}\left[\mathbf{g}(\bm{\theta}_k, \bm{u}, r)\right] - \nabla \mathcal{L}(\bm{\theta}_k) \right\|.
\end{equation}

{\setlength{\abovedisplayskip}{4pt}
 \setlength{\belowdisplayskip}{4pt}
\begin{align}
b 
&= \left\| \mathbb{E} \left[ \frac{1}{2\epsilon} \left( \mathcal{L}(\bm{\theta}_k + \epsilon r \bm{u}) - \mathcal{L}(\bm{\theta}_k - \epsilon r \bm{u}) \right) K(r) \bm{u} \right] - \nabla \mathcal{L}(\bm{\theta}_k) \right\| \notag \\
&\overset{\text{\circled{1}}}{=} \left\| \mathbb{E} \left[ \frac{1}{2\epsilon} \left( \mathcal{L}(\bm{\theta}_k + \epsilon r \bm{u}) K(r) \bm{u}  \right) - \nabla \mathcal{L}(\bm{\theta}_k) \right] \right\| \notag \\
&\overset{\text{\circled{2}}}{=} \left\| \mathbb{E} \left[ \frac{1}{2\epsilon} \mathcal{L}(\bm{\theta}_k + \epsilon r \bm{u}) K(r) \bm{u} \right] - \nabla \mathcal{L}(\bm{\theta}_k) \right\|  \notag \\
&\overset{\text{\circled{3}}}{=} \left\| \mathbb{E} \left[ (\nabla \mathcal{L}(\bm{\theta}_k + \epsilon r \bm{u})r K(r)) - \nabla \mathcal{L}(\bm{\theta}_k) \right] \right\| \notag \\
&\overset{\text{\circled{4}}}{\leq} \sup_{z \in S_2^d(1)} \mathbb{E} \left[ \left( \nabla_z \mathcal{L}(\bm{\theta}_k + \epsilon r \bm{u}) - \nabla_z \mathcal{L}(\bm{\theta}_k) \right) r K(r) \right]  \notag \\
&\overset{\text{\circled{5}}}{\leq} \kappa_\beta \epsilon^{\beta - 1} \frac{L}{(l - 1)!} \mathbb{E}\left[\|\bm{a}\|^{\beta - 1}\right] 
\leq \kappa_\beta \epsilon^{\beta - 1} \frac{L}{(l - 1)!} \frac{d}{ d+\beta-1}  
\lesssim \kappa_\beta L \epsilon^{\beta - 1} \label{eq:bias_bound}
\end{align}
}

where $\bm{a} \in B_2^d(1)$, $\circled{1}$: distribution of $\bm{u}$ is symmetric; 
$\circled{2}$: directly simplify the equation;
$\circled{3}$: a version of the Stokes’ theorem~\cite{zorich2016mathematical} ; 
$\circled{4}$: gradient norm represent the supremum of directional derivatives 
$\nabla_z \mathcal{L}(\bm{\theta}) = \lim_{\epsilon \to 0} \frac{\mathcal{L}(\bm{\theta} + \epsilon z) - \mathcal{L}(\bm{\theta})}{\epsilon}$; $\circled{5}$ Taylor expansion.

(2) For the second moment of gradient approximation, it can be denoted as  
$\mathbb{E}\|\mathbf{g}(\bm{\theta}^*, \bm{u}, r)\|^2$. We have

{\setlength{\abovedisplayskip}{4pt}
 \setlength{\belowdisplayskip}{4pt}
 \vspace{-0.1in}
\begin{align}
\zeta^2 &=\mathbb{E}\|\mathbf{g}(\bm{\theta}^*, \bm{u}, r)\|^2= \mathbb{E}\left[\left\| \frac{1}{2\epsilon} \left( \mathcal{L}(\bm{\theta}^* + \epsilon r \bm{u}) - \mathcal{L}(\bm{\theta}^* - \epsilon r \bm{u}) \right) K(r) \bm{u} \right\|^2 \right] \notag \\
&= \frac{1}{4\epsilon^2} \mathbb{E} \left[ \left( \mathcal{L}(\bm{\theta}^* + \epsilon r \bm{u}) - \mathcal{L}(\bm{\theta}^* - \epsilon r \bm{u}) \right)^2 (K(r))^2 \right] \notag \\
&\overset{\circled{1}}{\leq} \frac{\kappa}{2\epsilon^2} \mathbb{E} \left[ \left( \mathcal{L}(\bm{\theta}^* + \epsilon r \bm{u}) - \mathcal{L}(\bm{\theta}^* - \epsilon r \bm{u}) \right)^2  \right] \notag \\
&\overset{\circled{2}}{\leq} \frac{\kappa}{2\epsilon^2} \left( \epsilon^2 \mathbb{E} \left[\left\| \nabla \mathcal{L}(\bm{\theta}^* + \epsilon r \bm{u}) + \nabla \mathcal{L}(\bm{\theta}^* - \epsilon r \bm{u}) \right\|^2 \right] \right) \notag \\
&= \frac{\kappa}{2\epsilon^2} \left( \epsilon^2 \mathbb{E} \left[\left\| \nabla \mathcal{L}(\bm{\theta}^* + \epsilon r \bm{u}) + \nabla \mathcal{L}(\bm{\theta}^* - \epsilon r \bm{u}) \pm 2\nabla \mathcal{L}(\bm{\theta}^*) \right\|^2 \right] \right) \notag \\
&\overset{\circled{3}}{\leq} 4 \left\| \nabla \mathcal{L}(\bm{\theta}^*) \right\|^2 + 4 \kappa L^2 \epsilon^2 \mathbb{E}[\|\bm{u}\|^2] \notag \\
&\overset{\circled{4}}{\leq} 4\kappa \sigma_*^2 + 4\kappa L^2 \epsilon^2 \mathbb{E}[\|\bm{u}\|^2]=4\kappa \sigma_*^2 + 4\kappa L^2 \epsilon^2 \label{eq:variance_bound}
\end{align}
}

where $\circled{1}$ Inequality of squared norm of the sum, inequality between positive random variables and independence of the noise; $\circled{2}$ Wirtinger–Poincaré inequality;
$\circled{3}$ $L$-smoothness of $\mathcal{L}$;
$\circled{4}$ Overparameterization assumption.

\textbf{Sketchup for covergence.}
According to ~\cite{woodworth2021even}, we can replace the following expression of the bound for biased oracle with the bias and second moment estimation above. And we have

{\setlength{\abovedisplayskip}{4pt}
 \setlength{\belowdisplayskip}{4pt}
 \vspace{-0.1in}
\begin{align}
\mathbb{E}[\mathcal{L}(\bm{\theta}_N^{ag}) - \mathcal{L}(\bm{\theta}^*)] 
&= c \left( \frac{L\rho^2}{N^2} + \frac{L\rho^2}{N} + \frac{\zeta \rho}{\sqrt{N}} + b\rho + \frac{b^2}{2L}N \right) \notag \\
&\lesssim \frac{L\rho^2}{N^2} + \frac{L\rho^2}{N} + \frac{1}{\sqrt{N}} 
\left( \sqrt{\kappa \sigma_*^2} + \sqrt{\kappa L^2 \epsilon^2}  \right)\rho \notag \\
&\quad + \left( \kappa_\beta L \epsilon^{\beta - 1}  \right)\rho 
+ \frac{1}{L} \left( \kappa_\beta L \epsilon^{\beta - 1} \right)^2 N \notag \\
&\leq \frac{L\rho^2}{N^2} + \frac{L\rho^2}{N} 
+ \frac{\sqrt{\kappa \sigma_*^2} \rho}{\sqrt{N}} + \frac{\sqrt{\kappa} L \epsilon \rho}{\sqrt{N}}  \notag \\
&\quad + \kappa_\beta L \epsilon^{\beta - 1} \rho 
+ \kappa_\beta^2 L \epsilon^{2(\beta - 1)} N \label{eq:final_bound}
\end{align}
}

We use
$\circled{1}$ $\frac{L\rho^2}{N^2}$,
$\circled{2}$ $\frac{L\rho^2}{N}$,
$\circled{3}$ $\frac{\sqrt{\kappa} \sigma_* \rho}{\sqrt{N}}$,
$\circled{4}$ $\frac{\sqrt{\kappa} L \epsilon \rho}{\sqrt{N}}$,
$\circled{5}$ $\kappa_\beta L \epsilon^{\beta - 1} \rho$,
$\circled{6}$ $\kappa_\beta^2 L \epsilon^{2(\beta - 1)} N$,
to denote every term in the expression.
Then we try to use $\psi$ to bound every term above and now we attempt to find all parameter values for all cases.
\begin{equation}
\mathbb{E}[\mathcal{L}(\bm{\theta}_N^{ag}) - \mathcal{L}(\bm{\theta}^*)]
= \frac{L\rho^2}{N^2} + \frac{L\rho^2}{N} + \frac{\sqrt{\kappa} \sigma_* \rho}{\sqrt{N}} 
+ \frac{\sqrt{\kappa} L \epsilon \rho}{\sqrt{N}}   
+ \kappa_\beta L \epsilon^{\beta - 1} \rho 
+ \kappa_\beta^2 L \epsilon^{2(\beta - 1)} N 
\leq6 \psi.
\end{equation}
For $\circled{1}$, $\circled{2}$, and $\circled{3}$, we have $N$:

{\setlength{\abovedisplayskip}{4pt}
 \setlength{\belowdisplayskip}{4pt}
 \vspace{-2mm}
\begin{align}
\frac{L \rho^2}{N^2} \leq \psi 
&\quad\Longrightarrow\quad
N \geq \sqrt{\frac{L \rho^2}{\psi}} = \mathcal{O}\left(\sqrt{\frac{L \rho^2}{\psi}}\right) \notag \\
\frac{L \rho^2}{N} \leq \psi 
&\quad\Longrightarrow\quad
N \geq \frac{L \rho^2}{\psi} \notag \\
\frac{\sqrt{\kappa} \sigma_* \rho}{\sqrt{N}} \leq \psi 
&\quad\Longrightarrow\quad
N \geq \frac{\kappa \sigma_*^2 \rho^2}{\psi^2} \label{eq:N_bounds}
\end{align}
}

\vspace{-1mm}
Note that $\frac{L \rho^2}{\psi} > \sqrt{\frac{L \rho^2}{\psi}}$. That leads to the following expression for $N$
\begin{equation}
N =T= \mathcal{O}\left( \max\left( \frac{L \rho^2}{\psi}, \frac{\kappa \sigma_*^2 \rho^2}{\psi^2} \right) \right).
\end{equation}
For $\circled{4}$, $\circled{5}$, and $\circled{6}$, we can have $\epsilon$:

{\setlength{\abovedisplayskip}{4pt}
 \setlength{\belowdisplayskip}{4pt}
 \vspace{-2mm}
\begin{align}
\frac{\sqrt{\kappa} L \epsilon \rho}{\sqrt{N}} &\leq \psi 
\quad \Longrightarrow \quad 
\epsilon \leq \frac{\sqrt{N} \psi}{\sqrt{\kappa} L \rho} 
= \max\left( \frac{\sqrt{L \rho^2 / \psi} \cdot \psi}{\sqrt{\kappa} L \rho}, 
              \frac{\sqrt{\kappa \sigma_*^2 \rho^2 / \psi^2} \cdot \psi}{\sqrt{\kappa} L \rho} \right) \notag \\
&= \max\left( \frac{\psi^{1/2}}{\sqrt{\kappa} L^{1/2}}, \frac{\sigma_*}{L} \right) \notag \\[1ex]
\kappa_\beta^2 L \epsilon^{2(\beta - 1)}N &\leq \psi 
\quad \Longrightarrow \quad 
\epsilon \leq \left( \frac{\psi}{\kappa_\beta^2 LN} \right)^{\frac{1}{2(\beta - 1)}} \notag \\[1ex]
\kappa_\beta L \epsilon^{\beta - 1} \rho &\leq \psi 
\quad \Longrightarrow \quad 
\epsilon \leq \left( \frac{\psi}{\kappa_\beta L \rho} \right)^{\frac{1}{\beta - 1}} \label{eq:epsilon_bounds}
\end{align}
}

\begin{equation}
\epsilon = \min\left(
\max\left( \frac{\psi^{1/2}}{\sqrt{\kappa} L^{1/2}}, \frac{\sigma_*}{L} \right),\quad 
\left( \frac{\psi}{\kappa_\beta^2 LN} \right)^{\frac{1}{2(\beta - 1)}},\quad
\left( \frac{\psi}{\kappa_\beta L \rho} \right)^{\frac{1}{\beta - 1}}
\right)=(\frac{\psi}{\kappa_\beta L \rho} )^{\frac{1}{\beta - 1}}.
\end{equation}

\end{document}